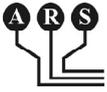

# Dynamic Balance Control of Multi-arm Free-Floating Space Robots

**Panfeng Huang**[1], **Yangsheng Xu**[1] **& Bin Liang**[1,2]
[1]Department of Automation and Computer-Aided Engineering
The Chinese University of Hong Kong, Hong Kong, P.R. China
[2]Shenzhen Space Technology Center
Harbin Institute of Technology, Shenzhen, P.R. China
pfhuang@acae.cuhk.edu.hk

*Abstract: This paper investigates the problem of the dynamic balance control of multi-arm free-floating space robot during capturing an active object in close proximity. The position and orientation of space base will be affected during the operation of space manipulator because of the dynamics coupling between the manipulator and space base. This dynamics coupling is unique characteristics of space robot system. Such a disturbance will produce a serious impact between the manipulator hand and the object. To ensure reliable and precise operation, we propose to develop a space robot system consisting of two arms, with one arm (mission arm) for accomplishing the capture mission, and the other one (balance arm) compensating for the disturbance of the base. We present the coordinated control concept for balance of the attitude of the base using the balance arm. The mission arm can move along the given trajectory to approach and capture the target with no considering the disturbance from the coupling of the base. We establish a relationship between the motion of two arm that can realize the zeros reaction to the base. The simulation studies verified the validity and efficiency of the proposed control method.*

## 1. Introduction

Unlike ground-base robot manipulator, the space manipulator has no fixed base. The dynamic reaction forces and moments due to the manipulator motion will disturb the space base, especially, when the space robot is in free-floating situation, The longer the motion time of space manipulator is, the greater the disturbance to the base will be. Hence, it is essential to resolve the attitude balance problem of a space robot during the manipulator operation. The position disturbance of the base may not be an issue. The attitude change, however, is more serious because the solar panel and tele-command from the ground station should keep their life lines. On the other hand, the attitude changes possibly cause the serious collision between the manipulator hand and the target. Therefore, the attitude balance problem of free floating space robot is a challenge problem about the application of space robot in space operation.

Nowadays, space robots are aiding to construct and maintain the International Space Station (ISS) and servicing the space telescope. Therefore, space robotic servicing for satellite such as rescue, repair, refuelling and maintenance, promising to extend satellite life and reduce costs, made it one of the most attractive areas of developing space technology. Space robots are likely to play more and more important roles in satellite service mission and space construction mission in the future. A well known space robot, the Shuttle Remote Manipulator System (SRMS or "Canadarm") [D. Zimpfer and P. Spehar, 1996] was operated to assist capture and berth the satellite from the shuttle by the astronaut. NASA missions STS-61, STS-82, and STS-103 repaired the Hubble Space Telescope by astronauts with the help of SRMS. Engineering Test Satellite VII (ETS-VII) from NASDA of Japan demonstrated the space manipulator to capture a cooperative satellite whose attitude is stabilized during the demonstration via tele-operation from the ground base controln station [I. Kawano, et al. 1998], [Oda, M.1999], [Oda, M., 2000], [Noriyasu Inaba, et al, 2000]. The space robotic technologies mentioned above demonstrated the usefulness of space robot for space service. During the ETS-VII operation, the engineers considered the coordinated control of the manipulator and space base in order to compensate the reaction forces and moments. Obviously, compensating the attitude disturbance of space base will consume much fuel during the motion, at the same time, the fuel is very limit and not reproduced. The life of space robot will be reduced by a long way. Therefore, the coordinated control of manipulator and space base is not the best method to control the space robot.

There are many studies on dynamic interaction of space base and its manipulators. Dubowsky and Torres[Steven Dubowsky, Miguel A. Torres, 1991] proposed a method called the Enhanced Disturbance Map to plan the space manipulator motions so that the disturbance to the space base is relatively minimized. Their technique also aided to understand the complex problem and developed the algorithm to reduce the disturbance. Evangelos G. Papadopoulos[Evangelos G. Papadopoulos, 1992] exhibited the Nonholonomic behavior of free-floating space manipulator by planing the path of space manipulator. They proposed a path planning method in cartesian space to avoid dynamically singular configurations. Yoshida and Hashizume [K.Yoshida, et, al, 2001] utilized the ETS-VII as an example to address a new concept called Zero Reaction Maneuver (ZRM) that is proven particularly useful to remove the velocity limit of manipulation due to the reaction constraint and the time loss due to wait for the attitude recovery. Moreover, they found that the existence of ZRM is very limited for a 6 DOF manipulator,besides the kinematically redundant arm. However, the redundant manipulator give rise to the computational cost and complicated kinematics problems. Hence, it is not best way to use the redundant manipulator from the view of engineering point.

Nearly all examples mentioned above almost focused on the nature characteristics of the space robot with a single robot arm. Moreover, these previous research largely limited the application of manipulator so that the space manipulator can not do desired mission like as ground-base manipulator. Thus, the space manipulator's motion must follow the constrained path in order to minimize the disturbance to the space base. Therefore, the manipulator can not dexterously be operated to accomplish the desired mission. i.e, if the manipulator tracks the desired mission trajectory, the disturbance to the base can not be minimized. Hence, it is very difficult to minimize the disturbance or ZRM for desired capture mission. Fortunately, an obvious characteristic of the space robot with respect to ground-base robot is the dynamics coupling between the robot and it's base. Therefore, the dynamics coupling provides some potential methods for researchers and engineers. Using the dynamics coupling, we will address a novel methodology to compensate the disturbance other than the attitude control system (ACS) in this paper. In general, the dynamics coupling may be a disadvantage for space robot system, many previous research work focus on how to eliminate it. On the other hand, the dynamics coupling may be a advantage if we use it correctly. Using the nonholonomic path planning method to control robot motion, the desired attitude of the base can be maintained as proposed by Fernandes, Gurvits, and Li [C. Fernandes, et, al, 1992] and Nakamura and Mukherjee [Robert E Roberson et, al 1988], and the methods will be best implemented to a system with a strong dynamics coupling. Therefore, it is necessary to fully understand the dynamics coupling, and try to eliminate it completely, rather than to minimize the coupling effect at the beginning. The measurement of dynamics coupling was firstly addressed by Xu [ Yangsheng Xu et, al, 1993].

In this article, we discuss the problem of dynamic balance control of a space robot when space manipulator follows the specified path to capture the object, rather than use ACS to compensate dynamics coupling. Therefore, we proposed to use a balance arm to compensate the dynamics coupling due to mission arm's motion, thus, our idea changes the disadvantage of dynamics coupling as advantaged factor. Accordingly, This paper is organized as follows. We formulate the dynamics equations of multi-arm free-floating space robot, based on linear and angular momentum conservation law and Lagrangian dynamics in Section II, In section III, we analyze and measure the dynamics coupling between the base and its multi-arm manipulators. Then, we address dynamic balance control algorithm of the base during the motion of manipulators in Section IV. In Section V, the computer simulation study and result are shown. Final section summarizes the whole paper and draw the conclusion.

## 2. Multi-Arm Space Robot System

*2.1 Modeling of Multi-arm Free-Floating Space Robot*
In this section, we describe the model of multi-arm space robot. These arms are installed on a base (spacecraft or satellite), one of these arms is called Mission Arm which is used to accomplish the space mission, such as, capturing and repairing, the others are the balance arms which are used to balance the reaction motion due to the motion of the mission arm. Motion planning and control of the balance arms are our key problems which will be introducedn in the following sections. Here, the space robot system is in free-floating situation during operation in order to avoid the serious collisions and save the fuel. Therefore, no external forces or momenta are exerted on the space robot system. The linear and angular momentum are conserved in this situation.

The authors assume a model of multi-arm space robot system which is composed of a base and multiple manipulator arms mounted on the base. Fig.1 shows a simple model of the space robot system with multi-arms, which is regarded as a free-flying serial $nm+1$ link system connected with $nm$ degrees of freedom (DOF) active joints. Here, in order to clarify the point at issue, we make the following assumpoing assumptions:

(a) The each installed manipulator consists of *n* links, and each joint has one rotational degree of freedom and is rate controlled. The whole system is composed of rigid bodies. Moreover, there are no mechanical restrictions and no external forces and torques for simplicity, so that linear and angular momentum is conservative, and equilibrium of forces and moments, strictly hold true during the operation;
(b) the kinematics and dynamics analysis during the motion is in the inertial coordinate system;
(c) the DOF of space robot system in inertial coordinate system is *nm+1*. The attitude of satellite has three DOFs, and the translation of the satellite has three DOFs. Here, *n* represents the number of robot arm and *m* represents the DOF number of each robot arm.

As shown in Fig.1, the dynamics model of space robot can be treated as a set of *nm+1* rigid bodies connected with *nm* joints that form a tree configuration, each manipulator joint numbers in series of 1 to *m*. We define two coordinate systems. One is the inertial coordinate system $\Sigma_I$ in the orbit, the other is the base coordinate system $\Sigma_B$ attached on the base body with its origin at the centroid of the base. The COM is the center of total system mass. We use three appropriate parameters such as Roll, Pitch, and Yaw to describe the attitude of the base.

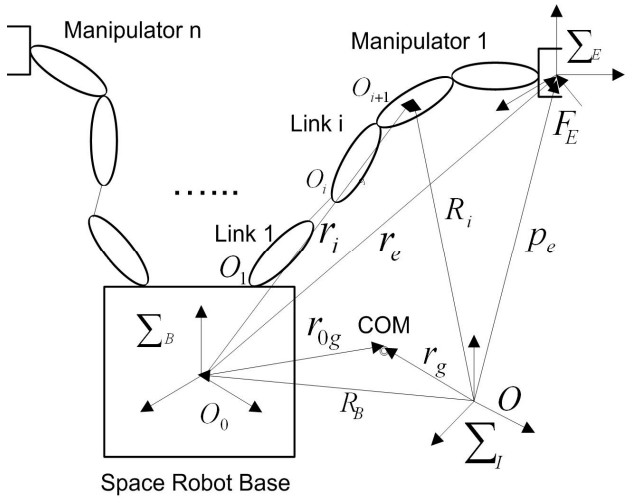

Fig.1. Dynamics Model of Space Robot System

*2.2 Equation of Motion*
In this section, the equation of motion of a multiple-arm free-floating space robot system are obtained according to proposed dynamics model. The dynamical influence due to orbitalmechanics are neglected because of relative short length and duration of the travel of system. It is a general problem to derive the dynamics equations of space robot system. We use Roberson-Wittenburg method [Robert E Roberson, et, al 1988] and previous work [E. Papadopoulos, et, al 1990] to derive the rigid dynamics of multi-body system. This method utilizes the mathematical graph theory to describe the interconnecting of the multi-body. The advantage of this method is that the various multi-body systems can be described by the uniform mathematical model. Thus, we will simplify describe the derivation process of dynamics equation according to the assumed model.The total kinetic energy of space robot system with respect to inertial coordinate $\Sigma_I$ can be expressed as follows.

$$T = T_0 + T_1 \qquad (1)$$

with

$$T_0 = \frac{1}{2} M v_{CM}^2 \qquad (2)$$

$$T_1 = \frac{1}{2}\{m_0 \dot{R}_B^2 + \Omega_0 I_0 \Omega_0\} + \frac{1}{2}\left\{\sum_{i=1}^{n}\sum_{j=1}^{k}\left(m_{(i,j)}\dot{r}_{(i,j)}^2 + \omega_{(i,j)} I_{(i,j)} \omega_{(i,j)}\right)\right\} \qquad (3)$$

where $I_{(i,j)}$ is the inertial matrix of *j*-th body of *i*-th manipulator arm with respect to system COM. Therefore, the kinetic energy of whole system can be rewritten as.

$$T = \frac{1}{2} v^T H_{(X_0, \theta)} v \qquad (4)$$

where $v = (\dot{X}_0^T, \dot{\theta}^T)$ is the vector of generalized velocities, and $X_0 = (X_0^T, \psi^T)$ is the vector to represent the position and orientation of the base, and matrix $H_{(X_0, \theta)}$ is an positive definite mass matrix, and

$$\theta = (\theta_1^1, \ldots, \theta_k^1, \ldots, \theta_1^n, \ldots, \theta_k^n)^T$$

From the kinetic energy formulation, we can derive dynamics equation by Lagrangian dynamics when there are not external forces or momenta exerted on the space robot system.

$$H^* \ddot{\phi} + C^*_{(\phi, \dot{\phi})} = \tau \qquad (5)$$

where $H^*$ is a generalized inertial matrix and $C^*$ is a centrifugal and coriolis term.

$$H^* = \begin{bmatrix} H_0 & H_{0m} \\ H_{0m}^T & H_m \end{bmatrix} \qquad (6)$$

$$C^* = \dot{H}^* \dot{\phi} - \frac{\partial}{\partial \phi}\left(\frac{1}{2}\dot{\phi}^T H^* \dot{\phi}\right) \qquad (7)$$

$$\phi = [X_0^T, \theta^T]^T \qquad (8)$$

The symbols in the above equations and Fig.1 are defined as follows:
$m_{(i,j)}$: mass of link *j* of the *i*-th manipulator arm

$r_{(i,j)}$: position vector of centroid of link $j$ of the $i$-th manipulator arm in $\Sigma_I$ system

$R_{(i,j)}$: position vector of joint $j$ of $i$-th manipulator arm in $\Sigma_B$ system

$p_{(i,e)}$: position vector of end-effector of $i$-th manipulator arm in coordinate system origin in $\Sigma_B$ system

$R_{(i,e)}$: position vector of end-effector of $i$-th manipulator arm in coordinate system origin in $\Sigma_B$ system

$R_B$: position vector of centroid of the base body in $\Sigma_I$ system

$r_g$: position vector of a total centroid of the space robot system in $\Sigma_I$ system

$I_{(i,j)}$: inertia tensor of link $j$ of $i$-th manipulator with respect to its mass center

$\tau$: joint torque of the whole manipulator arms

$\phi$: joint variables and base variables

$x_0$: position/orientation of the base

$H_0$: the inertial matrix of the base

$H_m$: the inertial matrix for the manipulator arms

$H_{0m}$: coupling inertial matrix between the base and robot manipulator

All vectors are described with respect to the inertial coordinate system $\Sigma_I$ and body coordinate system $\Sigma_B$. In Equation (5), $C^*$ can be obtained by inverse dynamics computation.

## 3. Dynamics Coupling

The dynamics coupling may not always be a disadvantage for the space robot system, as it can be used to balance the attitude of the base sometimes. In this paper, we will introduce the balance concept to compensate the attitude disturbance using dynamics coupling. This section addresses the characteristics and measure method of dynamics coupling.

The dynamics coupling is significant in understanding the relationship between the robot joint motion and the resultant base motion, and it is useful in minimizing fuel consumption for base attitude control. The measure of dynamics coupling was proposed by Xu[Yangsheng Xu, 1993]. Here, we will simply derive the mapping relationship between manipulator motion and the base motion for multi-arm space robot system.

Let $V_{(i,j)}$ and $\Omega_{(i,j)}$ be linear and angular velocities of the $j$-th body of the $i$-th manipulator arm with respect to, $\Sigma_I$ and let $v_{(i,j)}$ and $\omega_{(i,j)}$ be that with respect to $\Sigma_B$. Thus, we can obtain.

$$V_{(i,j)} = v_{(i,j)} + V_0 + \Omega_0 \times r_{(i,j)}$$
$$\Omega_{(i,j)} = \omega_{(i,j)} + \Omega_0 \tag{9}$$

where $V_0$ and $\Omega_0$ are linear and angular velocities of the centroid of the base with respect to $\Sigma_I$. The operator $\times$ represents the outer product of $R^3$ vector. the velocities $v_{(i,j)}$ and $\omega_{(i,j)}$ in the base coordinate $\Sigma_B$ can be calculated by following equation.

$$\begin{bmatrix} v_{(i,j)} \\ \omega_{(i,j)} \end{bmatrix} = J_{(i,j)}(\theta)\dot{\theta} \tag{10}$$

$$J_{(i,j)} = \begin{bmatrix} J_{T(i,j)} \\ J_{R(i,j)} \end{bmatrix} \tag{11}$$

where $J_{(i,j)}(\theta)$ is the jacobian of the $j$-th body of the $i$-th manipulator arm.

Therefore, the total linear and angular momentum of the space robot system P, L is expressed as follows.

$$P = m_0 V_0 + \sum_{i=1}^{n}\sum_{j=1}^{k}\left(m_{(i,j)} m_{(i,j)}\right) \tag{12}$$

$$L = I_0 \Omega_0 + m_0 R_0 \times V_0 +$$
$$\sum_{i=1}^{n}\sum_{j=1}^{k}\left(I_{(i,j)}\omega_{(i,j)} + m_{(i,j)}r_{(i,j)} \times v_{(i,j)}\right) \tag{13}$$

These equations are rewritten with the variables $v_0, \omega_0$ and $\dot{\theta}$.

$$\begin{bmatrix} P \\ L \end{bmatrix} = H_0 \begin{bmatrix} V_0 \\ \Omega_0 \end{bmatrix} + H_m \dot{\theta} \tag{14}$$

where

$$H_0 = \begin{bmatrix} \omega E & -\omega[r_{0g} \times] \\ \omega[r_g \times] & I_\omega \end{bmatrix} \tag{15}$$

$$H_m = \sum_{i=1}^{n}\sum_{j=1}^{k}\left(J_{R(i,j)}^T I_{(i,j)} J_{R(i,j)} + m_{(i,j)} J_{T(i,j)}^T J_{T(i,j)}\right)$$

$$J_{R(i,j)} = \begin{bmatrix} 0,\ldots,0, d_{(1,j)}, d_{(2,j)}, \ldots, d_{(i,j)}, 0, \ldots 0 \end{bmatrix} \tag{17}$$

$$J_{T(i,j)} = \begin{bmatrix} 0,\ldots,0, d_{(i,j)} \times (r_{(i,j)} - d_{(i,j)}), 0, \ldots 0 \end{bmatrix} \tag{18}$$

$J_{R(i,j)}$ is Jacobian matrix of $j$-th joint of $i$-th manipulator, and map the angular velocities of joint and angular velocity of $j$-th link. As such, the $J_{T(i,j)}$ maps the linear velocity. $E$ is unit vector indicating a rotational axis of joint $j$ of manipulator $i$. $r_{(i,j)}$ is position vector of the mass center of link $j$ of manipulator $i$, $d_{(i,j)}$ is position vector of joint $j$ of manipulator $i$.

In the free-floating situation, no external forces or momenta exerted on the whole space robot system, and we ignore the gravitational force. Therefore, the linear and angular momenta of the system are conserved. we assume that the initial state of system is stationary, so that the total linear and angular momenta are zero. Hence, Equation (14) can be rewritten as follows.

$$\begin{bmatrix} V_0 \\ \Omega_0 \end{bmatrix} = -H_m^{-1} H_B \dot{\theta} \quad (19)$$

Equation (19) describes the mapping relationship between the manipulator joint motion and the base motion. Because the velocities of end-effector in base coordinate is related to the joint velocities, we can obtain.

$$\begin{bmatrix} v_{(i,e)} \\ \omega_{(i,e)} \end{bmatrix} = J_{(i,j)} \dot{\theta} \quad (20)$$

However, the velocity of end-effector in inertial coordinate $\Sigma_I$ can be represented as follows.

$$V_{(i,j)} = v_{(i,j)} + V_0 + \Omega_0 \times r_{(i,e)}$$
$$\Omega_{(i,e)} = \omega_{(i,e)} + \Omega_0 \quad (21)$$

Combining Equation (19), (20), and (21), the relationship between the velocities of end-effector and the base velocities in inertial coordinate can be obtained as follows.

$$\begin{bmatrix} V_{(i,j)} \\ \Omega_{(i,j)} \end{bmatrix} = \left( J_{(i,j)} - K H_m^{-1} H_0 \right) \begin{bmatrix} V_0 \\ \Omega_0 \end{bmatrix} \quad (22)$$

where

$$K = \begin{bmatrix} E & [R_{(i,e)} \times] \\ 0_{3\times 3} & E \end{bmatrix} \quad (23)$$

In Equation (22), let $M = J_{(i,j)} - K H_m^{-1} H_B$, and the inverse relation of Equation (22) is

$$\begin{bmatrix} V_0 \\ \Omega_0 \end{bmatrix} = N \begin{bmatrix} V_e \\ \Omega_e \end{bmatrix} \quad (24)$$

The matrix M and N describe how robot motion alters the base motion, and these two matrices represent the dynamics interaction between the robot manipulator and the base, They are known dynamics coupling factor[14]. Equation (24) defines the interactive effect between the end-effector's motion of manipulator and the base motion. The dynamics coupling factor defined above basically represents the coupling of motion. Similarly, we also can define the coupling factor for force transmission from the end-effector to the base, thus we can easily obtain the force relationship as follows.

$$F_B = N^{-T} F_e \quad (25)$$

During the capture operation of space manipulator, the impact between the end-effector and the target is not avoided, so that, Equation (25) can measure the disturbance force to the base due to the collision. On the other hand, these mapping relationship can be used to plan carefully the robot manipulator motion in order to minimize the disturbance to the base.

In fact, the dynamics coupling of space robot system is very important to understand the dynamics characteristics and design the control part of the system. Especially, in the free-floating situation, this coupling must be considered in order to realize the minimum disturbance. In the next section, we will design the balance control algorithm to compensate the disturbance using dynamics coupling factor mentioned above.

**4. Dynamic Balance Control**

*4.1. Dynamic Balance Control Concept*
In this section, we address a balance control strategy for multi-arm space robot system. Here, we assume that the space robot system has a two-arm robot for simplicity, The number of DOF of each arm is same or not. We define these two arms as the mission arm and the balance arm, respectively. The mission arm is used to accomplish desired mission and the balance arm is used to compensate the attitude of the base using the dynamics coupling concept.
In order to balance the attitude disturbance of the base well, the authors utilize the coordinated control concept[Yangsheng Xu, et, al 1994] to address the dynamic balance control method proposed by them. However, according to our research problems, we can not use the attitude control system to compensate the robot arm's reaction so
that we use the balance robot arm to compensate the disturbance. Therefore, we address our balance control concept as follows.
(a) For two-arm space robot system, since the DOF of the whole system becomes so large that it is difficult to handle computation and commands using the state-of-the-art onboard computer. Hence, we suggest that each robot arm's controllers are independent.
(b) Because the motion trajectory of the mission arm has been planned for the desired operation, the reaction momenta to the base can be predicted using the dynamics coupling factors in advance.
(c) We build the mapping relationship of the mission arm's motion and balance arm's motion in the zero attitude disturbance condition. Thus, the trajectory of the balance arm will be defined by the trajectory of the mission arm.
The balance control concept proposed by us uses the motion of the balance arm to compensate the attitude disturbance due to the motion of the mission arm. we will address the balance control algorithm in detail in the following subsection.

*4.2 Coordinated Control of Two Arms*
Coordinated control of space robot was addressed by Oda, M.,[M. Oda, 1996], he presented the coordinated control concept between the spacecraft and its manipulator, Here, we use this concept to our problem.

we only consider the attitude disturbance of the base for free-floating space robot system. Hence, the angular momenta of the whole system can be estimated by Equation (13). At the beginning of the mission ram motion, the whole system keeps stable so that the initial momenta are zeros. Therefore, during the capture operation of the mission arm, we can get the relationship between the joint velocities of mission arm and the attitude angular velocities of the base from Equation (14) as follows.

$$\Omega_0 = J_{(M,\omega)} \dot{\theta}_M \quad (26)$$

Differentiate Equation (26) with respect to time, we can obtain.

$$\dot{\Omega}_0 = \dot{J}_{(M,\omega)} \dot{\theta}_M + J_{(i,j)} \ddot{\theta}_M \quad (27)$$

Where

$$J_{(M,\omega)} = -\widetilde{H}_{(M,m)}^{-1} \widetilde{H}_B \quad (28)$$

$$\widetilde{H}_B = I_{(M,\omega)} + W\widetilde{r}_g \widetilde{r}_{0g} \quad (29)$$

$$\widetilde{H}_{(M,m)} = I_{(M,\theta)} - \widetilde{r}_g J_{T\omega} \quad (30)$$

However, the angular momentum given by Equation (26) can not be integrated to obtain the base orientation $\psi$ as a function of the system's configuration, $\theta_M$ because of the nonholonomic characteristic, i.e we can not obtain the analytical solution to depict the relationship between the attitude $\psi$ and joint angle $\theta_M$. However, Equation (26) can be integrated numerically, thus, the orientation of the base will be a function of the trajectory of manipulator in joint space, i.e. different trajectory in joint space, with the same initial and final points, will generate different orientation of the base. The non-integrability property is known nonholonomic characteristics of free-floating space robot[E. Papadopouls, 1992].

For the given capture operation, the mission arm motions along the given trajectory. According to the trajectory planning method and dynamics equations, we can get the joint angle, angular velocities, and angular acceleration of the mission arm in joint space. Therefore, we can obtain the angular velocities, and acceleration of the base using Equations (26), (27) as mentioned above. On the other hand, we also can numerically integrate Equation (26) to obtain the orientation function of the base with respect to the joint variable of the mission arm. Therefore, we can obtain the torque exerted on the base because of the motion of the balance arm. Generally speaking, the attitude control system drives the reaction wheels to counteract this torque in order to balance the attitude of the base. our concept is to utilize the balance arm to substitute for the reaction wheels. an important advantage of this method is to save the fuel that is limit and valuable in space.

Hence, we use the motion parameters of the base due to the mission arm's motion to calculate the joint angle, joint angular velocities and joint acceleration of the balance arm. These parameters of the balance arm consist of the balance arm's trajectory in joint space. To realize the dynamic balance of the base's attitude, the attitude changes of base due to balance arm's motion $\varphi$ should satisfy $\psi + \varphi = 0$, so that the attitude disturbance of the base almost be zero. Thus, We can obtain the following two equations to represent the relationship between the balance arm and base's attitude.

$$\dot{\theta}_B = J_{(B,\omega)}^{-1} \Omega_0 \quad (31)$$

$$\ddot{\theta}_B = \dot{\Omega}_0 - \dot{J}_{(B,\omega)} \dot{\theta}_B \quad (32)$$

where $\theta_B$ is the joint variables of balance arm, and $J_{(B,\omega)}$ is the Jacobian matrix of balance arm.

For numerical simulation, Equation (31) can be integrated numerically to obtain the joint angle trajectory of the balance arm. Thus, we can get the desired trajectory of the balance arm in joint space, the balance arm's motioning along this desired trajectory can dynamically balance the attitude of the base. Thereby, our balance control concept can be realized. The effectiveness of balance control algorithm can be demonstrated in the simulation study of the following section.

## 5. Simulation Study

To better understand and examine the practical balance control performance of the proposed method, we define a two-arm space robot system as an illustrative example to attest the balance control method. In simulation, we assume that the object is in the workspace of the mission arm for simplicity. We have carried out relatively precise numerical simulation with a realistic 3D model.

The model comprises a two-arm manipulator which is called mission arm and balance arm, respectively. Each arm has 3-DOF, and every joint is revolution joint. Moreover, the geometric structure of two-arm manipulators is same in order to simplify the computation.

|  | Space Base (Link 0) | Space Manipulators | | | | | |
|---|---|---|---|---|---|---|---|
|  |  | Mission Arm | | | Balance Arm | | |
|  |  | Link1 | Link2 | Link3 | Link1 | Link2 | Link3 |
| Mass ($Kg$) | 300 | 5.0 | 5.0 | 5.5 | 5.0 | 5.0 | 5.5 |
| Length ($m$) | 1.0 | 0.5 | 0.5 | 0.5 | 0.5 | 0.5 | 0.5 |
| $I_{xx}$ ($Kg \cdot m^2$) | 50 | 0.4187 | 0.004 | 0.0044 | 0.4187 | 0.004 | 0.0044 |
| $I_{yy}$ ($Kg \cdot m^2$) | 50 | 0.4187 | 0.1062 | 0.1168 | 0.4187 | 0.1062 | 0.1168 |
| $I_{zz}$ ($Kg \cdot m^2$) | 50 | 0.16 | 0.4187 | 0.4606 | 0.16 | 0.4187 | 0.4606 |

Table 1. Parameters of Space Robot System

Here, we assume that each link of manipulator is cylindric configuration.

The radius of link $r = 0.04m$, and the base of space robot system is cubic configuration. Table I shows the dynamics parameters of the two-arm manipulator space robot system.

In simulation, we firstly plan the mission arm trajectory using Point to Point algorithm (PTP) so that the mission arm can approach and capture the target. However, the orientation of end-effector of the mission arm will be effected because the dynamics coupling in free-floating situation so that the collision between the end-effector and target becomes very serious. Therefore, the dynamic balance control concept can compensate this disturbance using the disturbance of motion of the balance manipulator. From the academic point of view, this method is very useful, the simulation result can verify our balance control algorithm. The balance control concept can be divided into two phases, Phase I mainly computes dynamics coupling and base motion trajectory because of disturbance of master manipulator motion in free-floating situation. Phase II calculates the required joint trajectory of the balance arm according to the estimation of Phase I. We will address these two phases in the following parts in detail.

Phase I: According to the planned joint trajectory of missionmanipulator, we design the PD feedback controller to drive the joint. We use the inverse dynamics to calculate the dynamics coupling due to the motion of mission manipulator, at the same time, we also can obtain the changes of orientation, angular velocities, and angular accelerative velocities of the base because of motion of the mission arm. Here, we use the precise numerical simulation so that we can integrate numerically Equation (26) to obtain the mapping relationship between the base orientation and the mission manipulator's configuration.

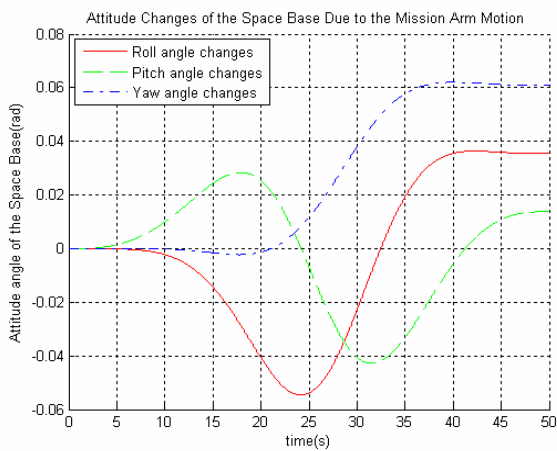

Fig. 2. Attitude changes of the base due to the mission arm motion

Phase II: We calculate the desired trajectory of the balance arm according to the motion trajectory of the base so that the disturbance of the balance arm's motion can compensate the disturbance of the mission arm's motion, i.e. combining the negative influence due to the mission arm's motion with the positive influence due to the balance arm's motion equals to zeros, Hence, the balance control method is realized. For the simulation result, Fig.2 and Fig.3 depict the attitude disturbance of the base because of the motion of the mission manipulator and that of the balance arm respectively.

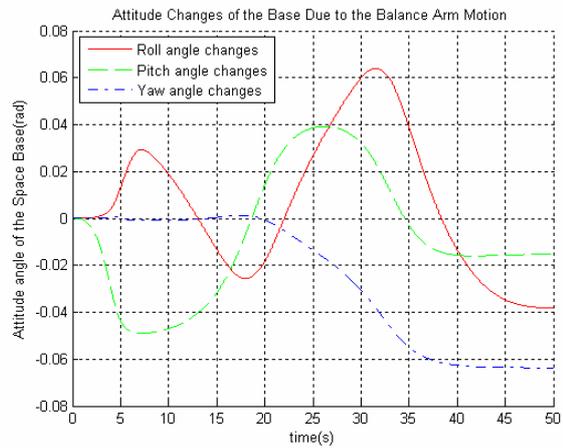

Fig. 3. Attitude changes of the base due to balance arm motion

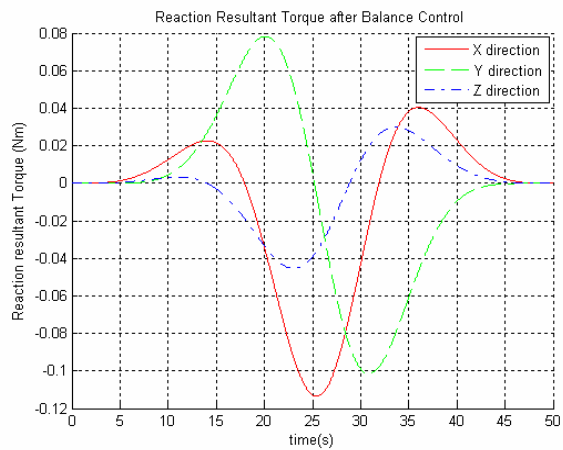

Fig. 4. Reaction resultant torque after balance control

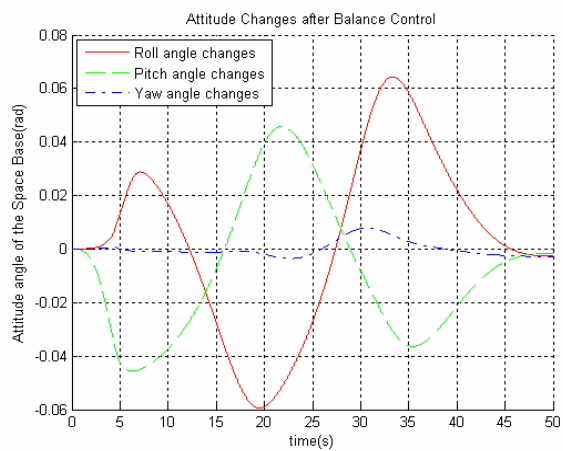

Fig. 5. Attitude changes after balance control

Fig.4 shows the reaction resultant torque to the base after using balance control algorithm.

Fig.5 depicts the attitude disturbance after using the balance control algorithm. Note that the attitude disturbance of the base becomes very small at the end of the capture operation, as shown in Fig.5. The goal of simulation is to verify that dynamic balance control method proposed is right and useful.

The authors also measure the dynamics coupling between the mission arm and the base in order to confirm whether the balance arm can balance this coupling and compensate the orientation disturbance of the base.

## 6. Conclusion

In this paper, we studied the dynamic balance control of multi-arm free-floating space robot. According to unique characteristics of free-floating space robot, we presented the dynamics coupling representing the robot arm and base motion and force dependency. Based on the dynamics coupling and measurement method, we proposed the concept of dynamic balance control, the use of the proposed concept is of significance in planing the balance arm's motion for compensate the attitude disturbance of space base.

Based on the dynamic balance control concept, we proposed the coordinated control algorithm for the mission arm and the balance arm. During the operation of mission arm, the balance arm can easily compensate the disturbance due to motion of the mission arm. The performance of the coordinated control algorithm is verified by the simulation studies. The simulation results showed that the proposed dynamic balance control method could be used practically.